\documentclass{article}

\pdfoutput=1

\PassOptionsToPackage{numbers, compress, sort}{natbib}
\usepackage[dblblindworkshop, final]{neurips_2026}
\workshoptitle{Dynamic Alignment in Human--AI Coupled System (BiAlign Workshop)}

\usepackage[utf8]{inputenc}
\usepackage[T1]{fontenc}
\usepackage{hyperref}
\usepackage{url}
\usepackage{booktabs}
\usepackage{amsfonts}
\usepackage{nicefrac}
\usepackage{microtype}
\usepackage{xcolor}

\usepackage[inline]{enumitem}

\AtBeginDocument{%
}

\newcommand{\cites}[1]{\begin{NoHyper}\citeauthor{#1}\end{NoHyper}'s~\citep{#1}}

\title{Trust Hinders Bidirectional Human--AI Alignment}

\author{%
  Michaela Benk\\
  Department of Informatics\\
  University of Zurich, Switzerland\\
  \texttt{michaela.benk@uzh.ch}
  \And
  Kacper Sokol\\
  Department of Informatics\\
  USI Lugano, Switzerland\\
  \texttt{kacper.sokol@usi.ch}
}

\begin{document}

\maketitle

\begin{abstract}
Trust calibration remains the dominant framing of AI-assisted decision making: the goal is to close the gap between perceived and actual trustworthiness of AI systems so that users rely on them appropriately. We argue that this approach, while useful for decision problems that are well-defined and clearly scoped, overly restricts what it means to make a decision with AI support. To overcome this limitation, we develop three dimensions of AI-assisted decision making. First, trusting AI is only one of several possible strategies for managing uncertainty. Second, trust calibration is often misapplied when AI models \emph{large-world} decision problems, i.e., ill-defined situations where goals and states are only partially known or unknowable. Third, in such settings human--AI alignment requires addressing higher-order decision processes. We conclude by reflecting on the implications of our observations on the design of process-oriented human--AI coupled systems that jointly identify and negotiate uncertainty throughout the decision-making task.
\end{abstract}

\section{Introduction}
\label{sec:introduction}

Imagine you are hiking across an unfamiliar terrain when you reach a fork in the trail.
Heavy fog obscures both paths;
a thunderstorm is approaching.
Conflicting reasons come to mind in favour and against each route.
In other words:
\emph{you are uncertain}.
You ask yourself which path is faster.
You also wonder whether choosing immediately is the right strategy.
Maybe you should rather wait for the fog to clear, seek higher ground, turn back or reach out to a more experienced hiker for advice.

Suppose that you instead consult a decision support tool (DST) that uses an artificial intelligence (AI) model trained on relevant hiking data. 
It recommends the left path, which is shorter, thus should take less time.
Assume that the suggestion is correct.
You follow it, but the path lacks any trees to shelter you from the rain.
The storm intensifies, you slip on wet ground and suffer a head injury.

What, if anything, does you trusting the DST tell us about the efficacy of your decision?
Was your choice good because the underlying recommendation was trustworthy?
Or was your decision bad because of the resulting head injury?
Would your partner or the medical professional treating you answer this question in the same way as the designer of the DST you used given the outcome?

Despite the richness of such decision scenarios, much of the literature on AI-assisted decision making evaluates this process primarily in terms of achieving well-calibrated trust and appropriate reliance~\citep{vereschakHowEvaluateTrust2021, wangHowAIAssistedDecisionMaking2026, ojewaleMoreNotBetter2026, mohleTrustMeBro2026}.
Yet in the vignette introduced above, the decision quality does not primarily depend on whether the AI model was trustworthy or the hiker exactly followed its advice.
Rather, it is determined by the joint alignment spanning multiple levels of uncertainty that the decision maker faces as well as the appropriate strategy to navigate it, which may or may not be to uncritically select a path that is recommended as the shortest.

The scenario described throughout this introduction aims to motivate our argument:
focusing primarily on trust confines AI assistance to an overly narrow conception of (human) decision making, which hinders the advancement of bidirectional human--AI alignment.
To substantiate this claim, we first overview relevant concepts (\autoref{prels}) and then develop three dimensions of AI-assisted decision making.
These span
\begin{enumerate*}[label=(\roman*)]
    \item
strategies to manage uncertainty (\autoref{claim1}),
    \item
characteristics of the decision problem (\autoref{claim2}) and
    \item
levels of higher-order decision processes (\autoref{claim3}).
\end{enumerate*}

Specifically, we show that trust calibration is only one of several possible approaches for managing uncertainty in decision making.
We argue that this particular strategy is useful for a confined class of problems, namely those characterised as \emph{small-world}, where the decision space and relevant states are well-defined and enumerable~\citep{gigerenzerPsychologicalAIDesigning2024}.
In \emph{large-world} problems, on the other hand, uncertainty may concern the decision space itself as well as high-level objectives, availability of alternatives and best strategies for their implementation.
In such settings, humans and AI tools may hold fundamentally different framings of the decision problem.
To achieve bidirectional human--AI alignment, we therefore propose shifting attention from trust calibration and reliance to understanding how humans and AI can jointly uncover, navigate and negotiate the uncertainty spanning different levels of AI assistance.
We finish with a discussion of limitations and counterarguments (\autoref{lims}) before concluding (\autoref{sec:conclusion}).

\section{Preliminaries and Definitions}
\label{prels}

Our three dimensions of AI-assisted decision making are built upon the notions of small- and large-world decision problems (\autoref{prels:sl}) and trust calibration (\autoref{prels:tc}).

\subsection{Flavours of Uncertainty in Small- and Large-world Decision Problems}
\label{prels:sl}

Decision theorists distinguish between situations of \textbf{risk}, \textbf{ambiguity} and \textbf{uncertainty}, accounting for how they affect information provision~\citep{savageFoundationsStatistics1972}.
The former two exist within small-world problems, whereas the latter one characterises large-world problems~\citep{gigerenzerPsychologicalAIDesigning2024}.

Formally, in a situation of \textbf{risk}, the set of all the future states together with the set of all their consequences are known.
Taking a risk -- or a chance -- is essentially a form of a gamble.
In contrast, when the probabilities of the consequences are unknown, but in principle knowable, it becomes a situation of \textbf{ambiguity}.
Further insights can thus close the information gap when dealing with both risk and ambiguity.
 
\textbf{Uncertainty}, on the other hand, refers to ill-defined settings.
Here, the future states and consequences -- or even the full account of the current situation -- are neither known nor knowable because this information cannot be obtained or changes at random.
Crucially, the Bayesian theory of expected utility maximisation applies \emph{only} to small-world problems.
\citet{savageFoundationsStatistics1972} went as far as considering it ``ridiculous'' to apply it in situations of uncertainty.

We return to how these different flavours of uncertainty apply to AI-assisted decision making in Section~\ref{claim2}.

\subsection{Origins, Definitions and Measurements of Trust Calibration}
\label{prels:tc}

The idea that trust can be calibrated to the right degree, such that appropriate reliance follows, originally arose in the trust-in-automation literature~\citep{benkTwentyfourYearsEmpirical2025, leeTrustAutomationDesigning2004, dzindoletRoleTrustAutomation2003}.
The central design problem therein was to help operators of automated systems -- deployed in work or industrial settings -- decide when to rely on their tools.
In this context, trust in an automated system is considered appropriate, calibrated or warranted when it is \emph{caused} by the trustworthiness of the system~\citep{jacoviFormalizingTrustArtificial2021}.

Although trust calibration and appropriate reliance are conceptually distinct, they are operationalised in a largely the same way~\citep{benkTwentyfourYearsEmpirical2025, leeTrustAutomationDesigning2004, dzindoletRoleTrustAutomation2003}.
For example, the number of times that a user agrees with and decides to follow AI predictions has been linked to both of these concepts~\citep{zhangEffectConfidenceExplanation2020, wuStepMINDVisualFramework2026, boRelyNotRely2025, chenMissingPiecesHow2025}.
A typical way to test whether trust in an AI tool is appropriate thus boils down to manipulating its trustworthiness and measuring the level of reliance exhibited by its users~\citep{jacoviFormalizingTrustArtificial2021}.
 
\section{Trusting AI Is One of Many Strategies for Managing Uncertainty}
\label{claim1}

\cites{millerExplainableAIDead2023} discussion of different reasoning processes involved in AI-assisted decision making hints at a crucial, yet frequently left implicit, distinction:
decisions come in \emph{layers}.
A \emph{first-order} decision concerns the choice itself;
for example, given a fork in a trail, which path should I take?
A \emph{second-order} decision concerns how the first-order choice is made;
for example, which strategy, rule or source of advice should I use when deciding?

\citet{sunsteinSecondOrderDecisions1999} provide an overview of key second-order strategies that people adopt:
rules, heuristics, presumptions, standards, routines, small steps, picking and delegating (see \autoref{tab:strategies} in \autoref{apx:strategies} for their description).
Several of these appear in the introductory vignette.
Such strategies are often chosen \emph{before} the first-order situation arises, precisely to reduce the cognitive burden of any eventual choice.
They work by restricting what the decision maker has to consider.

In \cites{sunsteinSecondOrderDecisions1999} terms, deciding whether to act on external advice, like an AI recommendation, is an instance of \emph{delegating}.
Consequently, trust calibration sits at a meta level:
it evaluates whether the delegate deserves the delegation.
Yet the literature mostly treats this higher-order concern as the primary decision problem, obscuring the process that characterises real-world decisions and hindering what human--AI alignment requires in such cases.

\section{Trust Calibration Is Misapplied to Large-world Decision Problems}
\label{claim2}

Since its origins in the trust-in-automation literature, the trust calibration paradigm has been stretched well beyond its intended setting, now covering medical diagnosis, law enforcement and financial investment~\citep{zhangEffectConfidenceExplanation2020};
most recently, it was applied to large language models supporting decisions across personal, legal, medical and scientific domains~\citep{ojewaleMoreNotBetter2026}.
What makes this problematic is that the underlying assumptions were adopted uncritically in these new contexts, where their suitability has not yet been examined or validated.

Trust calibration is defined as the correspondence between perceived and actual trustworthiness~\citep{jacoviFormalizingTrustArtificial2021}, which presupposes access to an oracle judging \emph{correctness} of AI outputs.
In the introductory vignette, the ground truth pertaining to which path is the shortest is known.
Yet this type of information remains unknowable for most large-world decisions~\citep{sokolArtificialIntelligenceShould2025}.
Returning to our hiking scenario, asking which path is the fastest or which one will not result in a head injury is ill-specified or unanswerable.

Trust calibration presupposes a \emph{stable} reference point for assessing trustworthiness;
in the hiking vignette, the measured distance of a path is unlikely to change.
In human--AI coupled systems, however, the AI may alter beliefs, preferences or goals of a user, thereby shifting the very benchmark against which its trustworthiness is judged.
As decision-aid scholars have long noted, preferences are rarely well-formed;
they rather emerge and evolve throughout the decision process itself~\citep{royEuropeanSchoolMCDA1996}.
Unsurprisingly, empirical studies tend to circumvent this limitation by imposing artificial stability.

Moreover, trust calibration presumes that the decision frame is \emph{fixed}, thus relevant states, consequences, objectives and alternatives are all known.
In this setting, trust calibration ``flattens'' the decision process to a single question:
whether to trust or not.
This narrow focus can nonetheless mask fundamental misalignments on other levels.
For example, if the hiker in our vignette follows a trustworthy AI recommendation and gets hurt, nothing in the interaction revealed whether the choices and strategies considered were appropriate.

Although trust is generally recognised as a dynamic construct~\citep{castelfranchiTrustTheorySocioCognitive2010}, trust calibration is measured through reliance on isolated recommendations, with its success evaluated by assessing joint performance.
This focus on \emph{outcomes} nonetheless obscures where the real difficulty lies when dealing with large-world decisions.
As highlighted by \citet{kerenAssessmentDecisionQuality2003}, once the structure and states of a problem are given, the decision itself is nearly trivial.
This is because the difficulty resides in structuring the problem space as well as reflecting on the possible states and the uncertainties or probabilities associated with different outcomes.

Consequently, many empirical studies inadvertently reformulate the large-world problems that they attempt to address as to satisfy methodological assumptions, effectively creating their ill-formed small-world variants.

\section{Human--AI Alignment in Large-world Settings Requires Addressing Higher-order Decision Processes}
\label{claim3}

We posit that addressing the challenges discussed thus far and achieving bidirectional human--AI alignment in decision making requires a shift in perspective.
From the decision maker's standpoint, delegating to a DST is one second-order strategy among many.
Choosing it presupposes that the (small- or large-world) decision problem as well as the structure and role of AI assistance are fixed;
as meaningful interaction becomes redundant in this setting, bidirectional alignment opportunities vanish.
To address this problem, we propose a rudimentary hierarchy that distinguishes several possible levels of AI assistance.
\begin{description}[nosep]
    \item[First-order: choice.]
    What is the best choice among the available options?
    \item[Second-order: strategy selection.]
    Which decision strategies are available and appropriate?
    \item[Third-order: decision structure.]
    What are the relevant goals, objectives, alternatives and criteria?
\end{description}

At the most basic level, a decision maker chooses among a set of alternatives.
When a DST proposes a single output and the user only decides whether to accept or reject it, the interaction follows the classical supervisory control paradigm from trust-in-automation research~\citep{leeTrustAutomationDesigning2004}.
In this setting, the AI effectively makes the choice and the human's role is reduced to approving or dismissing it.

Trust calibration operates as a proxy one tier above this choice, helping the user to determine whether to rely on the AI in the first place.
AI support can also enter at higher levels in a collaborative manner:
it can help to structure the decision, identify appropriate decision strategies and determine whether delegating to AI is appropriate.
Defining these levels enables us to situate trust calibration, as it is typically operationalised, within a broader hierarchy of decision-making processes.
It further allows us to identify at what level uncertainty enters the picture and opens up a new space of user interaction.
The latter could, for instance, allow humans and AI to negotiate types and objects of uncertainty.

\section{Limitations and Counterarguments}
\label{lims}

Moving beyond outcome-based evaluation is inherently difficult.
A process-oriented approach judges decision quality by how well the decision structure represents the decision maker's goals and whether an appropriate strategy was chosen~\citep{kerenAssessmentDecisionQuality2003}.
Yet calling the hiker's choice ``successful'' after sustaining a head injury may strike most people as absurd, regardless of whether the underlying process was sound.
This psychological pull towards outcome evaluation thus impedes the shift we propose.

Another critique of our proposal is that trust calibration may simply require a better formalisation.
\citet{guoDecisionTheoreticFramework2024}, for instance, ground reliance in statistical decision theory, assuming known utilities, rational Bayesian updating and an optimal benchmark.
But these are precisely the small-world assumptions we argue against:
their framework works well because, and only when, the decision problem is already settled.
\citet{hullmanUnderspecifiedHumanDecision2025} support our view, showing that decision problems in AI-assisted decision-making studies are often underspecified.
However, by suggesting that studies not adopting a utility-theoretic framework cannot provide sufficiently rigorous normative claims and should therefore be regarded as primarily descriptive, their conclusion anchors the field to a particular conception of decision making that rules out precisely the settings of most interest.

Crucially, our work criticises the use of trust calibration as a general-purpose paradigm in AI-assisted decision making.
We neither argue that trust is irrelevant nor that it is the only focus of the field.
Our claim is that the field uses it beyond its scope without examining whether its assumptions still hold.

\section{Conclusion}
\label{sec:conclusion}

AI-assisted decision making has been primarily positioned as a problem of trust calibration, allowing to identify when a DST is trustworthy and when people should rely on its recommendations.
In this paper we argued that this framing is valuable for well-specified, small-world decision problems, but it becomes limiting when applied to large-world settings that increasingly characterise the practical use of AI.
Rather than treating the calibration of human trust in a DST as a form of alignment, we situate it within a process in which humans and AI tools jointly uncover, contest and negotiate the uncertainties that shape different levels of decision making.
We contend that these considerations are especially important in view of the recent shift toward developing AI for critical thinking and reasoning support~\citep{millerExplainableAIDead2023, shenPositionBidirectionalHumanAI2025,sokolArtificialIntelligenceShould2025}.

\bibliographystyle{ACM-Reference-Format}
\bibliography{biblio}

\clearpage\newpage

\appendix

\section{Overview of Second-order Strategies}
\label{apx:strategies}

\autoref{tab:strategies} provides an overview of second-order strategies -- introduced by \citet{sunsteinSecondOrderDecisions1999} -- that people adopt to manage uncertainty.

\begin{table*}[h]
\centering
\caption{
Second-order strategies for managing uncertainty according to \citet{sunsteinSecondOrderDecisions1999}.
Note that relying on an AI recommendation is an instance of \emph{delegating} -- see the last entry in the table.
}
\label{tab:strategies}
\small
\begin{tabular}{@{}lp{.84\textwidth}@{}}
\toprule
\textbf{Strategy} & \textbf{Description} \\
\midrule
Rules &
Adopt a rule in anticipation of challenging or repetitive decisions; for example, ``when it rains, always take the fastest route''.
A rule is a full, or nearly full, \emph{ex ante} specification of results in individual cases.
Rules are error-prone in a characteristic way:
they are over- and under-inclusive relative to the reasons that justify them. \\
\addlinespace
Heuristics &
Rely on a mental shortcut that bypasses case-by-case choices. \\
\addlinespace
Presumptions &
Act on the assumption that $P$ is true until circumstances $A$ arise, in which case stop relying on $P$ and reconsider the situation. \\
\addlinespace
Standards &
Adopt a vague, value-based criterion that structures judgment without settling it;
for example, drive ``responsibly'', act ``reasonably'' and convict only ``beyond a reasonable doubt''. \\
\addlinespace
Routines &
Rely on a behavioural pattern that is habit-like but more voluntary and self-conscious than a habit;
for example, lock the office door on every exit or take the same commute daily.
Occasional error is accepted as the price of not deciding. \\
\addlinespace
Small steps &
Make a narrow, reversible decision now and defer answering the larger question. \\
\addlinespace
Picking &
Decide on an arbitrary or random basis rather than choosing for a reason;
for example, flip a coin, letting an irrelevant factor be decisive. \\
\addlinespace
Delegating &
Hand the decision to another party, like a spouse, a colleague or an AI tool.
Delegation ranges from mere consultation all the way to fully relinquished control. \\
\bottomrule
\end{tabular}
\end{table*}

\end{document}